\documentclass{article}

\usepackage{PRIMEarxiv}

\usepackage[utf8]{inputenc} 
\usepackage[T1]{fontenc}    
\usepackage{hyperref}       
\usepackage{url}            
\usepackage{booktabs}       
\usepackage{amsfonts}       
\usepackage{nicefrac}       
\usepackage{microtype}      
\usepackage{fancyhdr}       
\usepackage{graphicx}       
\usepackage{xcolor}
\usepackage{amsmath}
\graphicspath{{media/}}     

\title{$\mathbf{\pi}$-{\bf \textsc{yalli}}: un nouveau corpus pour le nahuatl 
\thanks{\textit{\underline{Citation}}: 
\textbf{Torres-Moreno et al. $\pi$-YALLI: un nouveau corpus pour le nahuatl. 
}} 
}

\author{
  Juan-Manuel Torres-Moreno, Juan-José Guzm\'an-Landa, Graham Ranger \\
  Laboratoire Informatique d'Avignon / ICTT, Université d'Avignon,
  Avignon (France)\\
  \texttt{\{juan-manuel.torres,juan-jose.guzman-landa,graham.ranger\}@univ-avignon.fr} \\
   \And
  Martha Lorena Avenda\~no Garrido, Miguel Figueroa-Saavedra, Ligia Quintana-Torres \\
  Facultad de Matem\'aticas \& IEE, Universidad Veracruzana,
  Xalapa, Veracruz (Mexique)\\
  \texttt{\{maravendano,migfigueroa,liquintana\}@uv.mx} \\
   \And
   Carlos-Emiliano Gonz\'alez-Gallardo \\
   LIFAT, Université François Rabelais à Tours \\
   Tours (France)\\
   \texttt{gonzalezgallardo@univ-tours.fr} \\
   \And
   Elvys Linhares Pontes\\
   Trading Central Labs, Trading Central, France\\
   \texttt{elvys.linharespontes@tradingcentral.com} \\
   \And
   Patricia Vel\'azquez Morales \\
   Avignon, France \\
   \texttt{patricia\_velazquez@yahoo.com} \\   
   \And
   Luis-Gil Moreno Jiménez\\
   Paris, France\\
   \texttt{Luis-Gil.moreno\_Jimenez@sorbonne-universite.fr}\\
}

\begin{document}
\maketitle

\begin{abstract}
Le projet NAHU² est une collaboration franco-mexicaine qui vise à construire le corpus $\pi$-\textsc{yalli} adapté à l'apprentissage automatique et qui permettra par la suite de développer des ressources informatiques pour la langue nahuatl \cite{NAHU2}. Le nahuatl est une langue qui dispose de peu de ressources informatiques, bien qu'il s'agisse d'une langue vivante parlée par environ 2 millions de personnes. Nous avons décidé de construire $\pi$-\textsc{yalli}, un corpus qui permettra de mener des recherches sur le nahuatl afin de développer des Modèles de Langue (ML) dynamiques ou pas, qui permettront à leur tour de développer des outils de Traitement Automatique des Langues (TAL) tels que : a) un unificateur de graphèmes, b) un segmenteur de mots, c) un analyseur grammatical POS, d) un résumeur automatique de texte basé sur le contenu \cite{torres14} ; et éventuellement, e) un traducteur nahuatl-français (probabiliste ou basé sur l'apprentissage).
\end{abstract}


\section{Introduction}

Le {\bf nahuatl}, {\bf nawatl} ou {\bf mexicano} (ou encore \textbf{nahuatlahtolli} en nahuatl) est une langue autochtone de la famille Uto-Nahua \cite{smith2002aztecs,parlons}, parlée par un grand nombre de personnes au Mexique et dans d'autres régions d'Amérique Centrale\footnote{\url{https://fr.wikipedia.org/wiki/nahuatl}}. 
C'est une langue parlée au Mésoamérique depuis le Vème siècle, étant la langue nationale la plus parlée au Mexique, après l'espagnol, avec 1.651.958 nahuaphones \cite{inegi2020censo} et plus de 2,5 millions de personnes dans la nahuaphonie ou monde nahuaphone.

\begin{figure}
  \centering
  \includegraphics[width=8cm]{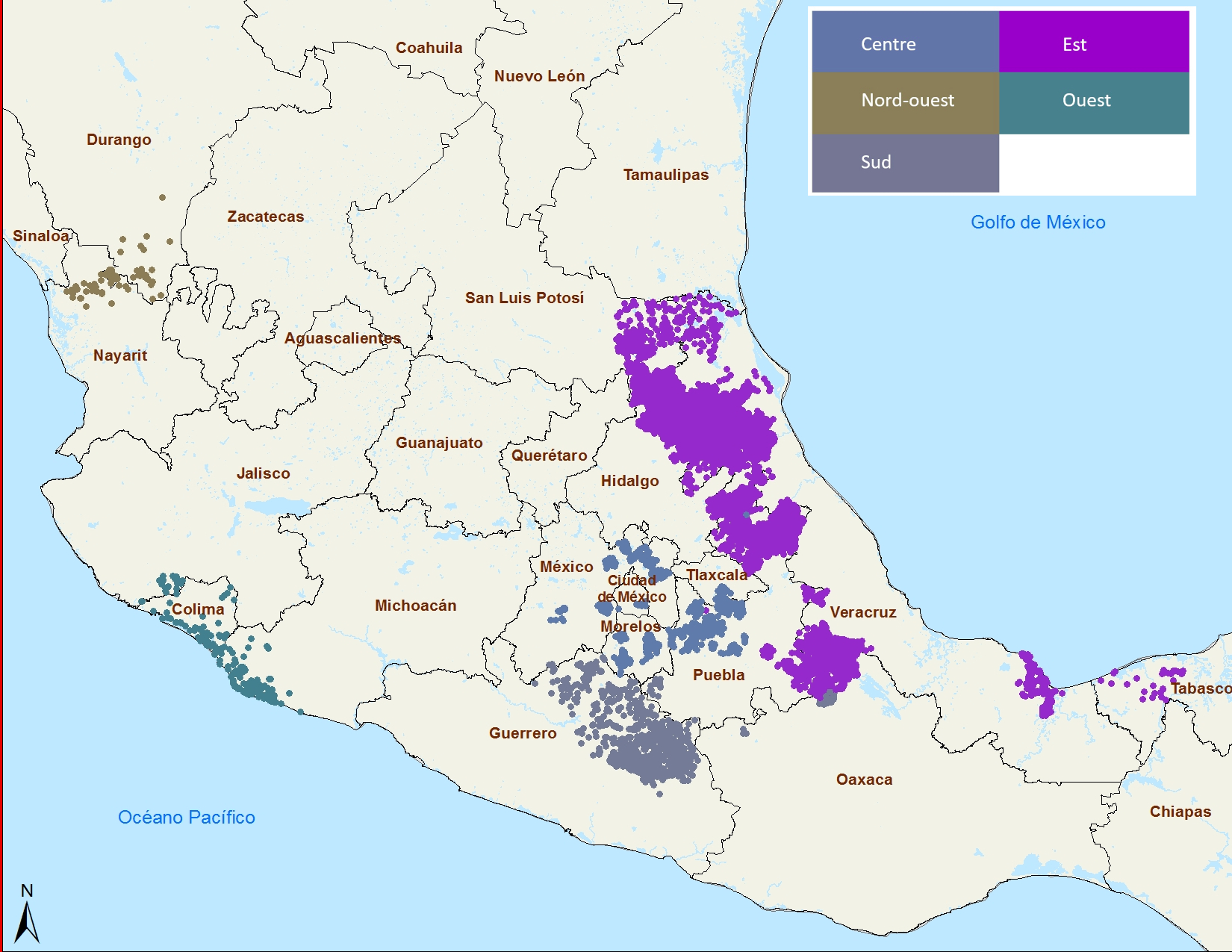}
  \caption{Principales variétés linguistiques du nahuatl parlées au Mexique.}
  \label{fig:nahuatl_variations}
\end{figure}

La Figure \ref{fig:nahuatl_variations} montre les principales variétés linguistiques de nahuatl parlées au Mexique selon les statistiques de l'INALI\footnote{\url{https://www.inali.gob.mx/clin-inali/html/l_nahuatl.html}}.
Des nos jours, le nahuatl est une langue dit vulnérable ou en danger de disparition, selon les variétés concernées, \cite{unesco2012atlas}. Ceci malgré les efforts continus que les communautés Nahua ont déployés depuis 2003 --année de la reconnaissance du nahuatl comme langue nationale-- pour que leur langue soit utilisée à l'oral et à l'écrit, dans l'industrie éditoriale, dans l'enseignement supérieur, les médias et les réseaux sociaux \cite{aguilar2023tecnologias, farfan2011proyecto, saavedra2023nawatlahtolli, olko2014toward}.

Le nahuatl est une langue polysynthétique et agglutinante, c'est-à-dire que les mots sont composés d'une racine verbale ou nominale et de morphèmes divers qui permettent de construire une signification. Un exemple d’agglutination est le suivant : 
\begin{itemize}
\item {\bf Cuauhtochtontli} (« \textit{Petit lapin sauvage} ») cuahuitl+tochtli+ton+tli (\textit{bois+lapin+diminutif+substantif})
\end{itemize}

Les relations syntaxiques entre mots sont établies par le biais de la valence du verbe et des connecteurs (particules). Ceux-ci peuvent également être composés par des regroupements établissant des nuances de sens, en plus des connecteurs discursifs. Certains de ces mots sont appelés mots-phrases, puisque leur morphologie inclut le sujet et le prédicat, en plus d'informations sur les actants, et d'éléments modaux, relationnels et directionnels. Un exemple de polysynthèse est : 
\begin{itemize}
\item {\bf Axcan timitzoncochtiah} (« \textit{Maintenant nous te faisons dormir loin} ») axcan ti+mitz+on+cochi+tia+h (\textit{maintenant nous+toi+directionnel+dormir+causatif+pluriel})
\end{itemize}

Malgré son riche héritage, de nos jours le nahuatl fait face à des défis importants en raison de son statut de langue minoritaire et de la pénurie de ressources informatiques disponibles pour sa préservation et sa diffusion. 
Or, plutôt de faire référence au nahuatl comme une \textit{langue minoritaire}, qui pourrait donner lieu à des interprétations biaisées, nous préférons considérer que le nahuatl fait partie de la catégorie des $\pi$-langues, ou langues \textit{peu dotées} de ressources informatiques \cite{these-pi} \footnote{Par opposition aux $\tau$-langues, ou langues {\it très bien} dotées de ressources informatiques et aux $\mu$-langues ou langues {\it moyennement} dotées de ressources informatiques.}.

\section{Corpus $\pi$-\textsc{yalli}}
\label{sec:headings}

On relève plusieurs efforts passés pour développer des corpus en nahuatl.
Un exemple en est le corpus \textit{Axolotl} \cite{gutierrez2016axolotl} disponible en ligne \footnote{\url{https://axolotl-corpus.mx}}, crée comme un corpus parallèle nahuatl-espagnol, et qui porte sur deux variétés de nahuatl. 
Cependant, des facteurs tels que le caractère oral de cette langue, le manque de standardisation des graphies ou encore le nombre important de variétés, font que l'on dispose de peu de documents.

Pour palier ce manque de ressources, nous avons décidé de développer un nouveau corpus pour le nahuatl. Nous avons collecté un ensemble de documents provenant de plusieurs sources et en différents formats (pdf, texte, doc/docx, odt, html, wiki) et encodages (iso-latin, us-ascii, utf8, utf16), ce qui a posé un certain nombre de problèmes dans le traitement informatique. 
La structure des documents étant hétérogène, nous les avons traités semi-automatiquement afin d'éliminer les informations non pertinentes. Ainsi les en-têtes, les indices, les tables, plusieurs références bibliographiques et, également des paragraphes écrits dans des langues autres que le nahuatl ont été supprimés des documents. 

Nous avons également eu accès à plusieurs sources documentaires numériques (dont quelques-unes qui sont confidentielles). 
Nous avons ainsi établi empiriquement les six catégories suivantes:
\begin{itemize}
    \item  documents historiques (HIS);
    \item  Wikipédia en nahuatl (WIK) \footnote{\url{https://nah.wikipedia.org}};
    \item  nouvelles, récits, poésie et contes (POE); 
    \item  documents juridiques et politiques (POL);    
    \item  mémoires de master, documents académiques divers, livres (LIV) et
    \item  publications scientifiques (SCI).
\end{itemize}

Nous avons aussi inclus les nombreux paragraphes de texte nahuatl appartenant au corpus libre Axolotl \cite{gutierrez2016axolotl}, issu du projet collaboratif \texttt{py-elotl} \footnote{\url{https://pypi.org/project/elotl/}}. 

Le corpus ainsi constitué $\pi$-\textsc{yalli} comprend environ 1.912M  \textit{tokens}\footnote{\textit{Tokens} dans le sens informatique du terme: des chaînes de caractères séparées par des espaces en blanc.} ou 14.8M de caractères (14.8 Mo de texte brut codé utf8) ayant le code ISO 639-3 \texttt{nah}\footnote{\url{https://iso639-3.sil.org/code/nah}}). 
Quelques statistiques basiques concernant la composition des catégories du corpus, sont présentées dans la Table (\ref{tab:statcorpus}). 

Les variétés incluses dans le corpus correspondent principalement aux variétés parlées dans l'État de Veracruz (nahuatl de la zone centrale et nahuatl de La Huasteca) également partagées avec d'autres États du centre et du nord du pays; et, de façon moins importante, à la variété nawat du sud de Veracruz et de l'État de Puebla, et à la variété \textit{tecpillahtolli } --un registre savant-- utilisée entre le XVIème et le XIXème siècles et qui a été employé dans les textes imprimés. Pour cette raison, des textes avec différentes graphies utilisées aujourd'hui et dans le passé ont été inclus.

\begin{table}[h!]
   \centering
   \begin{tabular}{|c|c|r|r|}
        \hline
        \bf Catégorie & \bf Documents & \bf Tokens & \bf \% Corpus \\ \hline
        HIS  & 2   & 2.468   & 0,13  \% \\ \hline
        WIK  & 1   & 194.275 & 10,16 \% \\ \hline
        POE  & 53  & 306.336 & 16,02 \% \\ \hline
        POL  & 13  & 327.206 & 17,10 \% \\ \hline
        LIV  & 81  & 394.287 & 20,61 \% \\ \hline
        SCI  & 16  & 688.270 & 35,98 \% \\ \hline
    \bf TOTAL&\bf 166 &\bf 1.912.842&\bf 100,00   \% \\ \hline

   \end{tabular}
   \caption{Statistiques basiques de documents du corpus $\pi$-\textsc{yalli} par catégorie. \label{tab:statcorpus}}
\end{table}

Une fois constitué, sous forme de texte brut avec métadonnées identifiant chaque texte, le corpus $\pi$-\textsc{yalli} sera mis en ligne\footnote{\url{http://juanmanueltorres.free.fr/corpus/piyalli/index.html}} pour une consultation par mots, suites de mots, regex, etc., par le biais de l'application CQPweb, interface graphique pour le Corpus Query Processeur \cite{evert2011twenty} (Evert \& Hardie, 2011). Dans un deuxième temps, il est prévu d'indexer une deuxième version du corpus, enrichie d'annotations grammaticales et de nouvelles métadonnées, dès que les outils seront disponibles. Cette version devra permettre des requêtes selon les lemmes ou les catégories grammaticales, en plus des critères déjà mentionnés.

\section{Modèles}
\label{sec:others}

Un Modèle de Langue (ML) est un outil computationnel conçu pour traiter et représenter les langues humaines. Au cœur de ces modèles réside l’utilisation des représentations vectorielles de mots, également appelées représentations denses, qui sont indispensables pour capturer les significations et les relations entre les mots dans un format adapté aux machines. Les représentations denses offrent un moyen puissant d’encoder à la fois des informations sémantiques et syntaxiques~\cite{almeida2023wordembeddingssurvey}. Ces représentations sont essentielles pour des applications nécessitant une compréhension sémantique avancée, telles que la reconnaissance des entités nommées, l’analyse de sentiments~\cite{linhares-pontes-etal-2018-predicting} et la classification et catégorisation automatiques des textes. 

Dans notre étude, nous nous concentrons initialement sur les Modèles de Langue statiques Word2Vec \cite{Mikolov2013distributed} et FastText~\cite{bojanowski-etal-2017-enriching}; par la suite nous allons utiliser un Grand Modèle de Langue léger (LLM, {\it Large Language Model}) du type BERT~\cite{devlin-etal-2019-bert}, tel qu'ALBERT, par exemple \footnote{\url{https://huggingface.co/docs/transformers/model_doc/albert}}. Word2Vec, avec ses architectures CBOW et Skip-Gram, capture les relations sémantiques basées sur les co-occurrences de mots dans de grands corpus, produisant ainsi des représentations vectorielles stables. FastText intègre des informations sur les sous-morphèmes\footnote{{\it Tokens} dans le sens d'apprentissage d'IA profonde du terme.}, ce qui le rend particulièrement efficace pour les langues morphologiquement riches, agglutinantes et pour le traitement de termes rares. 

Les modèles BERT génèrent de plongements des mots ou \textit{embeddings} dynamiques et contextuels, en tenant compte des nuances syntaxiques et sémantiques des phrases entières. Cependant, ces modèles ont besoin d'une grande quantité de données textuelles pour être performants. C'est une étude approfondie que nous mènerons avec des modèles de type BERT afin de construire le LM {\bf BERTL} ({\it BERt en nahuaTL}) et de pouvoir ainsi mesurer l'impact de la taille sur l'apprentissage et la performance dans des tâches spécifiques.

Chaque modèle ayant ses spécificités, la performance varie selon la langue, le domaine, les nuances sémantiques et la taille du corpus étudié. Nous évaluerons donc ces modèles pour le nahuatl afin de mesurer leur capacité à produire des représentations précises et cohérentes (intrinsèque). Ces évaluations ne se limiteront pas à des mesures quantitatives, mais incluront également des applications en aval pour mieux comprendre l'efficacité du modèle dans des tâches concrètes (extrinsèque). Les applications en aval incluront des tâches telles que la classification de texte, la classification des sentiments, la reconnaissance des entités nommées, le resumé automatique et la traduction automatique.

\section{Évaluation}

Nous avons établi un protocole de similitude sémantique pour réaliser une première évaluation de la qualité du corpus $\pi$-\textsc{yalli}.
Étant donné 23 termes de référence, chacun ayant associé une liste de 5 termes candidats, il a été demandé à 27 nahuaphones\footnote{Tous les annotateurs possèdent un niveau d'études universitaire.} de trier sémantiquement la liste de termes candidats, du plus proche au plus éloigné de la référence \cite{NAHU2}. 
Chaque terme candidat a reçu une note de 1 à 5 (étant 1 jugé le terme candidat le plus proche sémantiquement à la référence et 5 le plus éloigné). Ceci a permis de créer un ensemble de rangs.

La sélection des termes (références et candidats) a été réalisée selon différents critères : d'abord, les mots d'usage courant exprimés dans trois catégories grammaticales (substantifs, verbes et particules) comprenant les noms d'ustensiles, d'aliments, de vêtements, de couleurs, de goûts, de qualités, de termes de parenté et de parties du corps.
Ensuite, nous avons considéré des actions quotidiennes --au moyen des verbes transitifs, intransitifs, verbes d'état et de mouvement fléchis en numéro et formes verbales--, les particules adverbiales de nature quantitative et locative spatio-temporelle, et enfin des expressions de salutation. 
Ces mots ont été exprimés dans différentes variétés dialectales, y compris des formes caractéristiques du nahuatl central, du nahuatl de La Huasteque et du nahuatl du sud, ainsi que dans des formes caractéristiques du nahuatl savant ou littéraire ({\it tecpillahtolli}), et en utilisant des différents alphabets employés par les locuteurs, mais avec une majorité de formes caractéristiques du nahuatl central, écrites avec un alphabet modernisé.

Dans certains cas, où il y avait variation formelle, l'association a été réalisée en fonction des aspects morphologique ou compositionnel. Dans d'autres cas, il est possible que des sens figurés, symboliques ou métaphoriques aient été reconnus à partir d'une lecture plutôt culturelle. On a donc d’autres types d’association logique qui s’éloignent de l’association sémantique. Par exemple, dans le cas de la référence {\bf noyollo}, « mon cœur », où une association avec des termes comme {\bf nomah}, « ma main » ou {\bf noyoliknih}, « mon ami du cœur », était attendue, il y a eu une forte préférence pour le terme {\bf yoli}, « vivre », tout simplement parce qu'il était associé à son origine étymologique ({\bf yol-}) et donc à un certain sens originel du mot {\bf yollotl}, « coeur » ou « ce qui a de la vivacité ». D'un autre côté, on pense parfois à partir d'une traduction espagnole; tel est le cas de {\bf nemi}, « habiter, marcher », qui a tendance à être associé à {\bf yoli}, « être vivant », plutôt qu'à {\bf chantia}, « y résider ». associé à des formes qui peuvent composer une phrase pleine de signification, comme associer {\bf tlahtolli}, « mot, récit, langue », à {\bf onikkak}, « je l'ai écouté », et composer {\bf onikkak tlahtolli}, « je ai écouté un récit », ou {\bf noyollo} avec {\bf paki}, « être heureux », pour construire {\bf noyollo paki}, «~mon cœur est heureux », une salutation habituelle. Ainsi, nous trouvons des réponses qui montrent une différence culturelle entre ce que nous comprenons comme une association sémantique logique de nos langues, une extension des significations à partir d'usages culturels, une interférence sémantique dans des contextes de bilinguisme.
Ce type d'association n'était pertinent que pour certains mots bien spécifiques.
Les termes de référence eux-mêmes n'avaient pas de signification complexe, mais ils nécessitaient un profil de locuteur familier avec d'autres variantes et graphies. Le profil des annotateurs a été établi en tenant compte de cette condition, puisqu'il s'agit d'universitaires et d'étudiants de Master qui utilisent la langue nahuatl à l'oral et à l'écrit dans le cadre de leurs activités professionnelles, de formation et de communication.

Afin de pouvoir évaluer quantitativement l’accord de l'ensemble d'annotateurs, nous avons utilisé les deux métriques suivantes: 

\begin{itemize}
\item Coefficient de concordance $W$ de Kendall;
\item Entropie $H$ de Shannon.
\end{itemize}

Finalement, pour évaluer la performance des Modèles de Langue (ML) dans la tâche sémantique proposée, nous avons d'abord, établi un consensus des rangs des annotateurs $CR$, puis nous avons calculé le coefficient de corrélation $\tau$ de Kendall entre $CR$ et le rang produit par chacun des ML.

\subsection{Coefficient $W$ de Kendall}

Le coefficient $W$ de Kendall permet d’évaluer la cohérence entre plus de deux classements, et représente une extension du coefficient de corrélation $\tau$ de Kendall, conçu spécifiquement pour mesurer le degré de concordance entre deux classements. Étant donné $n$ éléments à classer et $m$ rangs indépendants (venant du même nombre de juges ou annotateurs) on calcule le coefficient de Kendall avec la formule suivante:

\begin{equation}
W=\frac{12\displaystyle\sum_{i=1}^{n}(R_i-\bar{R})^2}{m^2(n^3-n)}
\label{for:kendall}
\end{equation}

\noindent où $R_i$ est la somme des rangs assignés à l'élément $i$ et $\bar{R}$ la moyenne de ces sommes.
Une valeur de $W=1$ indique une concordance parfaite où tous les classements sont identiques, et une valeur de $W=0$ indique une absence totale de concordance, ce qui signifie que les positions dans les classements sont complètement incohérentes.

\subsection{Entropie $H$ de Shannon}
Le coefficient d'entropie de Shannon est utilisé pour déterminer l'incertitude ou hétérogénéité d'un ensemble.
Dans un ensemble de rangs, il renseigne sur la cohérence ou la diversité entre les rangs.
La statistique d'entropie de Shannon a été calculée en utilisant la formule suivante:

\begin{equation}
H(x)=-\displaystyle\sum_{i=1}^n p_i\log(p_i)
\label{for:entropia}
\end{equation}

\noindent où $p_i$ est la probabilité (ou fréquence relative) que l'élément $x$ soit dans la position $i$.
Les valeurs sont entre 0 (cohérence totale entre les rangs) et la valeur maximale possible $H_{\textrm{max}}=\log_2(n)$ (divergence totale entre les rangs qui a lieu quand toutes les positions sont équiprobables, c'est-à-dire $\frac{1}{n}$).

Nous avons normalisé la sortie de la métrique de Shannon entre $[0,\log_2(n)]$, où $n=5$ correspond au nombre de candidats par référence de la tâche sémantique. Dans le cas de Kendall nous avons employé directement sa valeur de sortie, car le coefficient $W$ est déjà normalisée entre [0,1].

\subsection{Exemple}

Nous présentons ci-après un exemple du protocole d'évaluation pour le terme de référence {\bf tototl} ($oiseau$). La table suivante résume cette information.
Soient les $n$ mots candidats $\{avion,aigle,coyote\}$ et les $m$ annotateurs $\{J_1,J_2,J_3\}$ qui ont annoté les rangs suivants:

Référence = {\bf tototl} ($oiseau$); Candidats = \{{\bf tepostototl} ($avion$), {\bf kwawtli} ($aigle$), {\bf koyotl} ($coyote$)\}
 
\begin{tabular}{l|c|c|c}
         & $avion$ & $aigle$ & $coyote$\\ \hline
         $J_1$  &  1&  2&  3 \\ \hline
         $J_2$  &  2&  1&  3 \\ \hline
         $J_3$  &  2&  1&  3 \\  \hline
\end{tabular}

Pour le calcul du coefficient $W$ de Kendall, on aura:

\begin{tabular}{c|c} 
          \hline
         $avion$ & $R_1=5$\\ \hline
         $aigle$ & $R_2=4$\\ \hline
         $coyote$ & $R_3=9$\\ \hline
         
\end{tabular}
~~et $\bar R=6$

En utilisant la formule (\ref{for:kendall}), avec $n=3$ et $m=3$ on obtient $W=0.778$

Pour le calcul de l'indice d'entropie de Shannon, nous avons calculé $H(x)$ pour chaque mot de référence. Pour le mot $avion$, on voit qu'il occupe une fois la position 1 ($p_1=\frac{1}{3}$) et deux fois la position 2 ($p_2=\frac{2}{3}$). En utilisant la formule (\ref{for:entropia}), on obtient pour ce mot $H(avion)\approx 0.918$. De la même manière, nous calculons cet indice pour les deux autres références, $H(aigle)\approx 0.918$ et $H(coyote)=0$. 
En fin, nous avons calculé la moyenne des trois entropies pour obtenir l'entropie totale de l'ensemble de rangs, $\bar H\approx 0.612$.
Dans le cas des deux mesures de l'exemple, $W$ et $\bar H$, on obtient plutôt un bon accord entre les trois annotateurs pour le terme de référence $oiseau$.

Maintenant, soient les rangs suivants:

Référence = {\bf tamalli} ($tamal$)\footnote{Le tamal [...] est un nom générique donné à plusieurs plats [...] d'origine indigène \url{https://fr.wikipedia.org/wiki/Tamal}}; Candidats = \{{\bf sakawilli} ($zacahuil$)\footnote{Une sorte de tamal mexicain, de grande dimension: \url{https://fr.wikipedia.org/wiki/Zacahuil}.}), {\bf tlaxkalli} ($tortilla$), {\bf nixtamalli} ($nixtamal$)\footnote{\url{https://fr.wikipedia.org/wiki/Nixtamalisation}} \}

\begin{tabular}{l|c|c|c}
         & $zacahuil$ & $tortilla$ & $nixtamal$\\ \hline
         $J_1$  &  1&  2&  3 \\ \hline
         $J_2$  &  2&  3&  1 \\ \hline
         $J_3$  &  3&  1&  2 \\  \hline
\end{tabular}

Pour le calcul de l'indice de Kendall on a:

\begin{tabular}{c|c} 
          \hline
         $zacahuil$ & $R_1=6$\\ \hline
         $tortilla$ & $R_2=6$\\ \hline
         $nixtamal$ & $R_3=6$\\ \hline
         
\end{tabular}
~~et $\bar R=6$

d'où $W=0$.

Dans le cas de l'entropie, nous avons $H(zacahuil)\approx H(tortilla)\approx H(nixtamal)=1.585$, donc l'entropie pour cet ensemble de termes est $H\approx 1.585$, ce qui correspond à la valeur maximale de $\log(3)$. Ainsi $W$ et $H$ impliquent un total désaccord des trois annotateurs concernant le terme $tamalli$.

\subsection{Évaluation des Modèles de Langue via un Rang par consensus}

En utilisant la méthode de Borda Count \footnote{\url{https://en.wikipedia.org/wiki/Borda_count}}, nous avons construit un Rang par consensus ($CR$), qui représente la majorité des rangs pour chacune des références. Puis, nous avons calculé la distance entre chacun des rangs des annotateurs et le rang par consensus $CR$. Ce consensus nous a servit principalement à deux choses: 
\begin{enumerate}
    \item  à estimer l'éloignement d'un annotateur donné par rapport à l'ensemble des $m$ annotateurs; et 
    \item à évaluer l'accord (via le coefficient de corrélation $\tau$ de Kendall) entre le consensus et les ML Word2Vec et FastText.
    \end{enumerate}
 
Il faut préciser que, dans le cas des annotateurs les plus éloignés du Rang consensus, nous avons décidé de les supprimer du protocole d'évaluation pour deux raisons: d'abord, car le rang induit par ces annotateurs peut constituer un biais au calcul des mesures, et ensuite, de manière plus importante, pour avoir un nombre impair (27) d'annotateurs, ce qui évite les potentiels ex-æquo entre les rangs. 

\section{Discussion}

\begin{figure}
  \centering
  \includegraphics[width=1\textwidth]{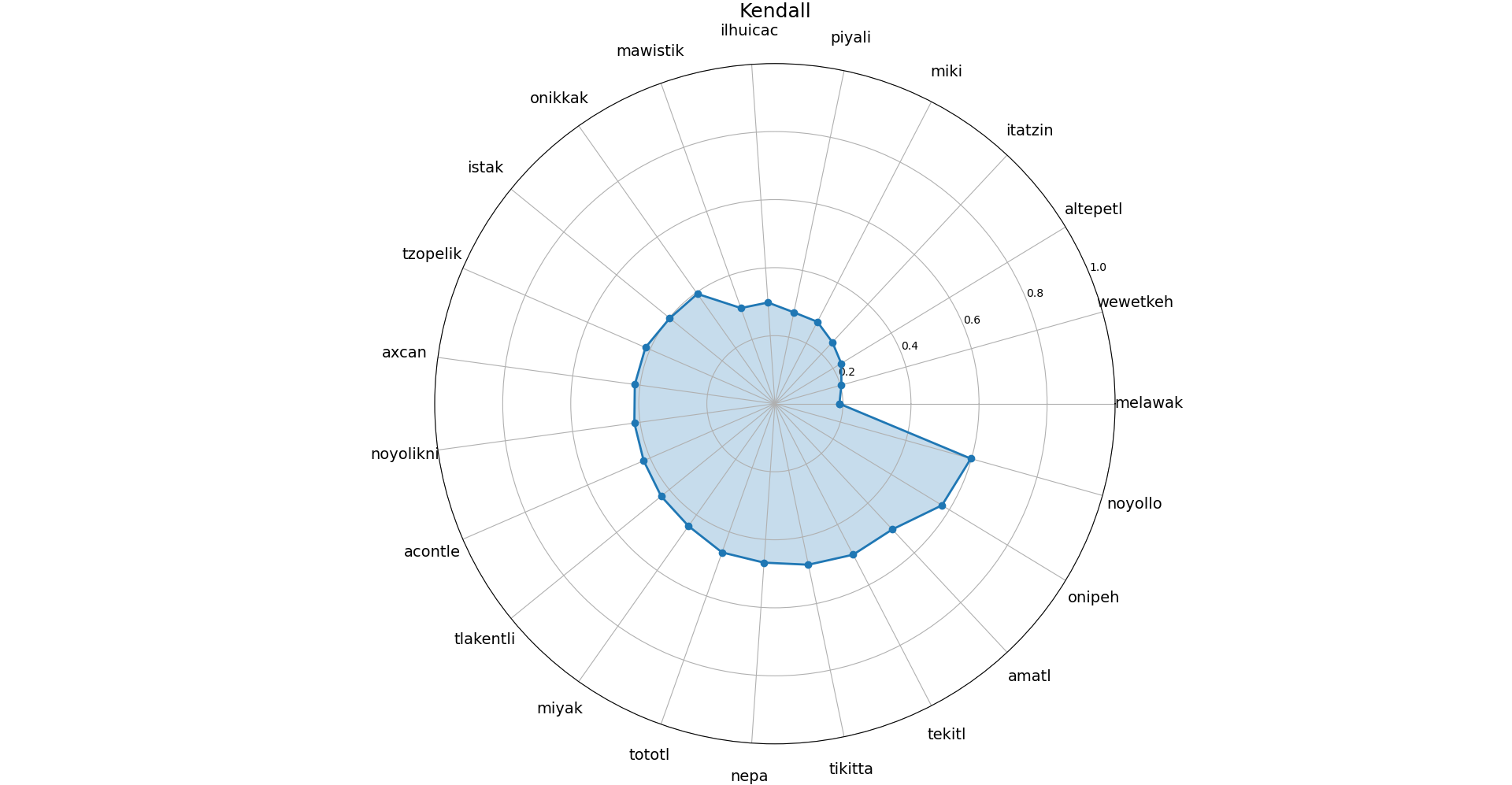}
  \caption{Coefficient $W$ de Kendall qui mesure l'accord entre les annotateurs, selon les termes de référence.}
  \label{fig:Kendall}
\end{figure}

\begin{figure}
  \centering
  \includegraphics[width=1\textwidth]{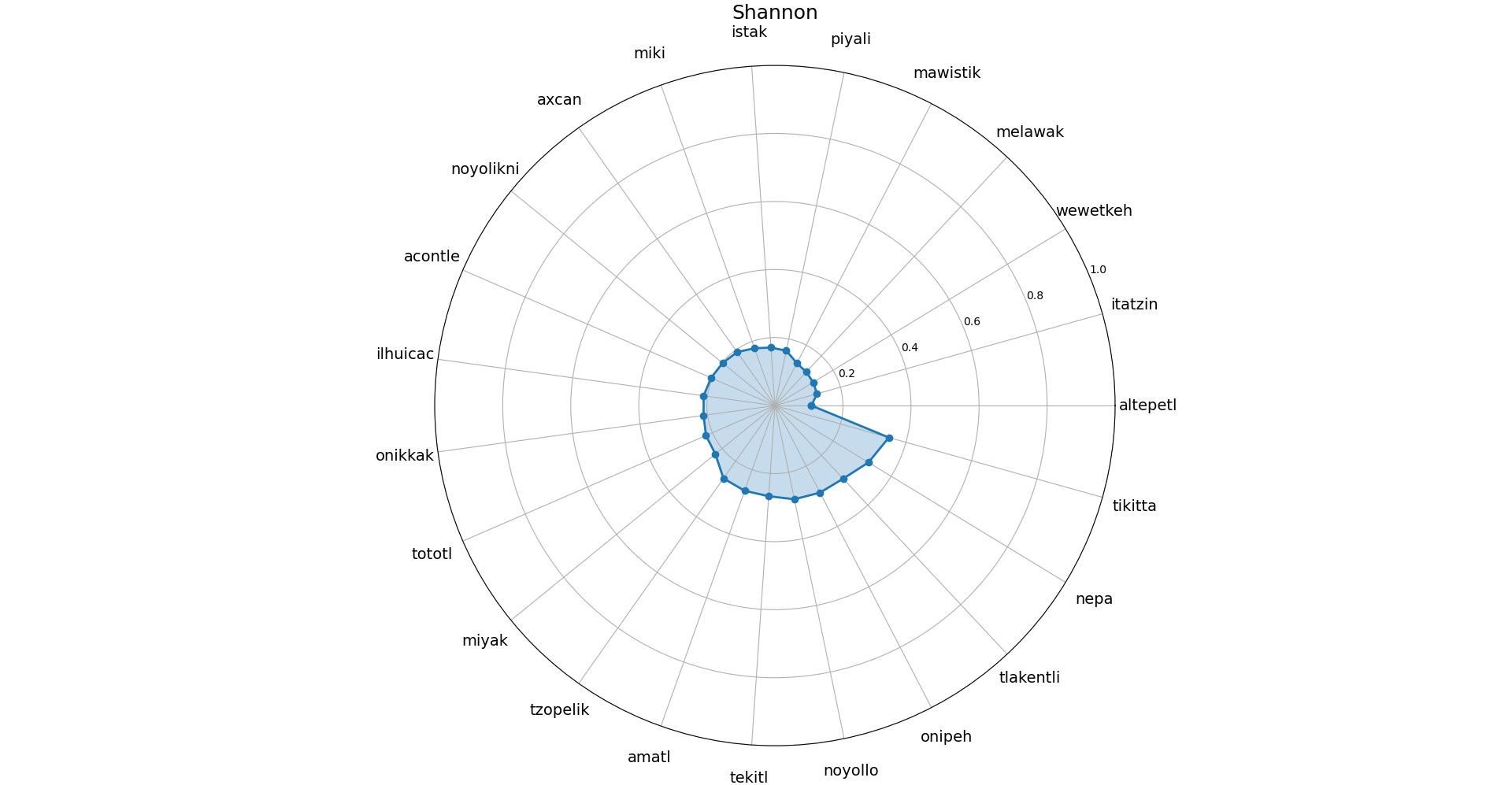}
  \caption{Métrique de Shannon $H$ normalisée, qui mesure l'accord entre les annotateurs selon les termes de référence.}
  \label{fig:Shannon}
\end{figure}

Nous avons constaté que les deux mesures que nous avons employées montrent une corrélation par rapport au --relativement-- faible accord entre les annotateurs (voir les Figures \ref{fig:Kendall} et \ref{fig:Shannon}).
Or, dans l'état de l'art, la métrique la plus adaptée et la plus utilisée pour mesurer l'accord entre plusieurs rangs, est la statistique $W$ de Kendall.
Nous avons donc focalisé sur cette métrique. 
Le coefficient $W$ de Kendall varie de 0,1896 pour la référence \textbf{Melawak} (\textit{correct}), qui a obtenu la valeur la plus basse, jusqu'à 0,598 pour la référence \textbf{noyollo} (\textit{mon c\oe ur}), avec la valeur la plus élevée.

D'après nos autres résultats, nous constatons que certaines références ayant des valeurs basses de Kendall (et également d'entropie) n'ont pas obtenu les réponses attendues.
En effet, quelques choix de classement montrent que certains mots candidats n'ont pas été clairement identifiés. 
Cela peut s'expliquer en raison de leur emploi très localisé qui a donc faussé les évaluations d'association sémantique. 
Il en va de même pour certains mots qui ont pu être considérés comme des archaïsmes et ou qui n'étaient pas connus. 
Dans d'autres cas, l'association semble avoir été faite sur la base d'une logique syntaxique, selon laquelle les verbes sont appariés avec des substantifs pouvant être objet ou sujet.

Des basses valeurs de Kendall et d'entropie témoigne peut-être des méconnaissances des graphies ou des mots peu utilisés parmi les diverses communautés nahuaphones, et en particulier pour les annotateurs ayant participé à la tâche.

La moyenne des coefficients de Kendall avec les 27 annotateurs, pour les 23 termes de référence est d'environ $\bar W =0,389$ et de $\bar H = 0,215$ pour l'entropie de Shannon, ce qui montre un certain degré de difficulté de cette tâche sémantique.

Enfin, le calcul des résultats des modèles Word2Vec, Glove et FastText est encore en cours, et nous devons raffiner le protocole présenté afin de focaliser sur les couples références-candidats les mieux adaptés au paramétrage des algorithmes. Nous proposerons également d'autres tâches TAL pour évaluer la qualité du corpus. Notamment nous utiliserons des modèles de type BERT pour établir des étalons plus adéquats d'évaluation.

\section{Conclusions}

Bien que le corpus $\pi$-\textsc{yalli} ait une taille réduite vis-à-vis de corpus d'autres langues, et qu'il soit encore en cours de développement, nous pensons qu'il s'agit d'une ressource intéressante pour étudier la langue nahuatl~\cite{NAHU2}.
On pourra, par exemple, étudier l'impact de la taille du corpus dans l'apprentissage --profond ou pas-- de Modèles de Langue nahuatl. 
Par ailleurs, nous augmentons constamment le volume du corpus $\pi$-\textsc{yalli}.
Ce corpus permettra de développer des outils d'analyse TAL classiques, des Modèles de Langue (ML) et probablement des Grands Modèles de Langue (LLM) d'Intelligence Artificielle légers que nous diffuserons à la communauté scientifique. De plus, l'utilisation émergente et croissante du nahuatl dans les réseaux sociaux, dans l'édition, dans les programmes universitaires et dans la diffusion scientifique rend ces outils de plus en plus nécessaires pour l'accès et la gestion de l'information numérique disponible. 

Ainsi, cette accessibilité permettra de relier différentes communautés de nahuaphones situées dans des régions et des pays différents, ainsi que de faire circuler les connaissances exprimées dans cette langue auprès des étudiants et des spécialistes. Il s'agit donc d'un élan puissant pour faire mieux connaître cette importante $\pi$-langue.

\section*{Remerciements}
Ces travaux ont été soutenus, en partie, par la Structure Fédérative de Recherche Agorantic, et d'un autré côté par l'École Universitaire de Recherche InterMEDIUS, pour le financement du projet NAHU² et la thèse de Juan-José Guzm\'an-Landa (LIA/Université d'Avignon) respectivement.

\bibliographystyle{unsrt}  
\bibliography{NAHU2Arxiv}  

\end{document}